\documentclass{Interspeech2024}
\usepackage{cite} 
\usepackage{dcolumn}
\usepackage{makecell}
\usepackage[ruled,linesnumbered]{algorithm2e}
\usepackage{algpseudocode}
\usepackage{mathtools}
\usepackage{microtype}
\usepackage{tikz}
\usetikzlibrary{shapes.geometric}

\newcolumntype{d}[1]{D{.}{.}{#1}}

\usepackage{color}

\newcommand{\data}[1]{\ensuremath{{#1}}}
\newcommand{\crv}[2]{{#1}\textbackslash{#2}}
\newcommand{\myparspace}{-3mm}
\newcommand{\ltg}{Liu  \textit{et al.}}
\newcommand{\repspace}{\ensuremath{\mathbb{R}^d}}

\interspeechcameraready

\title{Orthogonality and isotropy of speaker and phonetic information \texorpdfstring{\\}{} in self-supervised speech representations}

\name{Mukhtar}{Mohamed}
\name{Oli Danyi}{Liu}
\name{Hao}{Tang}
\name{Sharon}{Goldwater}

\address{
  University of Edinburgh, United Kingdom
  }
\email{mukhtaralgezoli@gmail.com, oli.liu@ed.ac.uk, hao.tang@ed.ac.uk, sgwater@inf.ed.ac.uk}


\begin{document}
\maketitle
 
\begin{abstract}
Self-supervised speech representations can hugely benefit downstream speech technologies, yet the properties that make them useful are still poorly understood.
Two candidate properties related to the geometry of the representation space have been hypothesized to correlate well with downstream tasks: (1) the degree of orthogonality between the subspaces spanned by the speaker centroids and phone centroids, and (2) the isotropy of the space, i.e., the degree to which all dimensions are effectively utilized.
To study them, we introduce a new measure, Cumulative Residual Variance (CRV), which can be used to assess both properties. 
Using linear classifiers for speaker and phone ID to probe the representations of six different self-supervised models and two untrained baselines, we ask whether either orthogonality or isotropy correlate with linear probing accuracy. 
We find that both measures correlate with phonetic probing accuracy, though our results on isotropy are more nuanced.
\end{abstract}

\vspace{-0.5em}

\section{Introduction}

Self-supervised speech representations have made a huge impact on downstream speech technologies, yet the properties that make their representations useful are still poorly understood.
Benchmarks indicate that both phone and speaker labels are, to a large degree, linearly separable in the representations of popular recent models \cite{yang21c_interspeech}, and beyond this, a number of studies have compared the extent to which these labels are recoverable from the representations of different models \cite{yang21c_interspeech,ma2021probing} or across different layers of the same model \cite{pasad2021layer,pasad2023comparative}. However, these analyses say little about {\em how} such information is represented, beyond just assessing the linear separability of classes. 
Here, we address this question using a {\em geometric} approach---an approach that is widely used for analyzing self-supervised models of text (e.g.,  \cite{mikolov2013linguistic,cai2020isotropy,chang2022geometry,hernandez2021low,mu2017all,timkey2021all}) 
as well as high-dimensional brain imaging data (e.g., \cite{kriegeskorte2013representational,saxena2019towards,sorscher2022neural}),
but has received only a little attention in the speech technology community \cite{stephenson2019untangling,abdullah21_interspeech,choi2022opening,liu23_orthogonal}. 

To assist our analysis, we develop a new measure for analyzing high-dimensional distributions, the Cumulative Residual Variance (CRV). When applied to datasets \data{X} and \data{Y} embedded in the same high dimensional space, the CRV of \data{X} with respect to \data{Y}, denoted \crv{\data{X}}{\data{Y}}, provides a quantitative measure of the degree to which the principal components of \data{Y} are orthogonal to those of \data{X}. Meanwhile,  \crv{\data{X}}{\data{X}} is a measure of the {\em isotropy} of \data{X}---the degree to which \data{X} effectively utilizes all dimensions of the embedding space, i.e., has uniform covariance \cite{rudman2022isoscore}.

Using this measure, we draw on two previous lines of work that suggest potentially fruitful analyses.  
First, we build on a recent study which analyzed LSTM models trained using two different loss functions and demonstrated that speaker and phonetic information were represented in orthogonal subspaces \cite{liu23_orthogonal}. 
The CRV measure allows us to better quantify orthogonality, and we use it
to analyze several additional models with a variety of architectures, loss functions, and training data. In experiments on English LibriSpeech, we show that, unlike randomly initialized (untrained) models, all trained models have a high degree of orthogonality between the speaker and phonetic subspaces.
In addition, for all six trained models, the accuracy of a phone classifier trained on the model representations is significantly correlated with the CRV between the two subspaces.

Next, we explore whether and how the isotropy of the representational space might predict phone or speaker classification accuracy. 
It has been argued in the NLP literature that higher isotropy is desirable in an embedding space (e.g., \cite{cai2020isotropy,mu2017all} and see review in \cite{rudman2023stable}). However, we did not find strong evidence for this hypothesis: when we computed the isotropy and phone (or speaker) classification accuracy for different layers of each model, we found a statistically significant correlation in only two out of six trained models. On the other hand, we did find a strong and consistent correlation between phone classification accuracy and the isotropy of the phone class centroids. This suggests that having evenly distributed centroids is more important for classification accuracy in these models than having evenly distributed frame representations.

\begin{figure*}[tb]
    \centering
    \includegraphics[width=8.8cm]{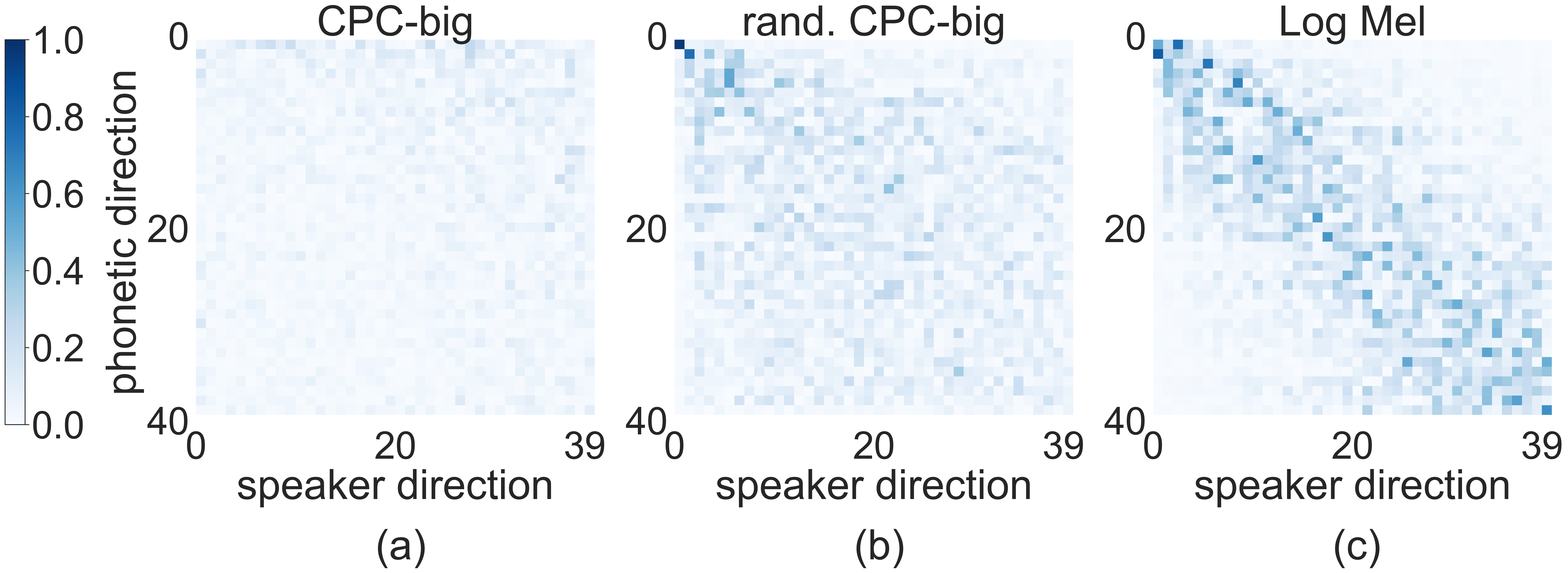}
    \hspace{0.5cm}
    \includegraphics[width=6.0cm]{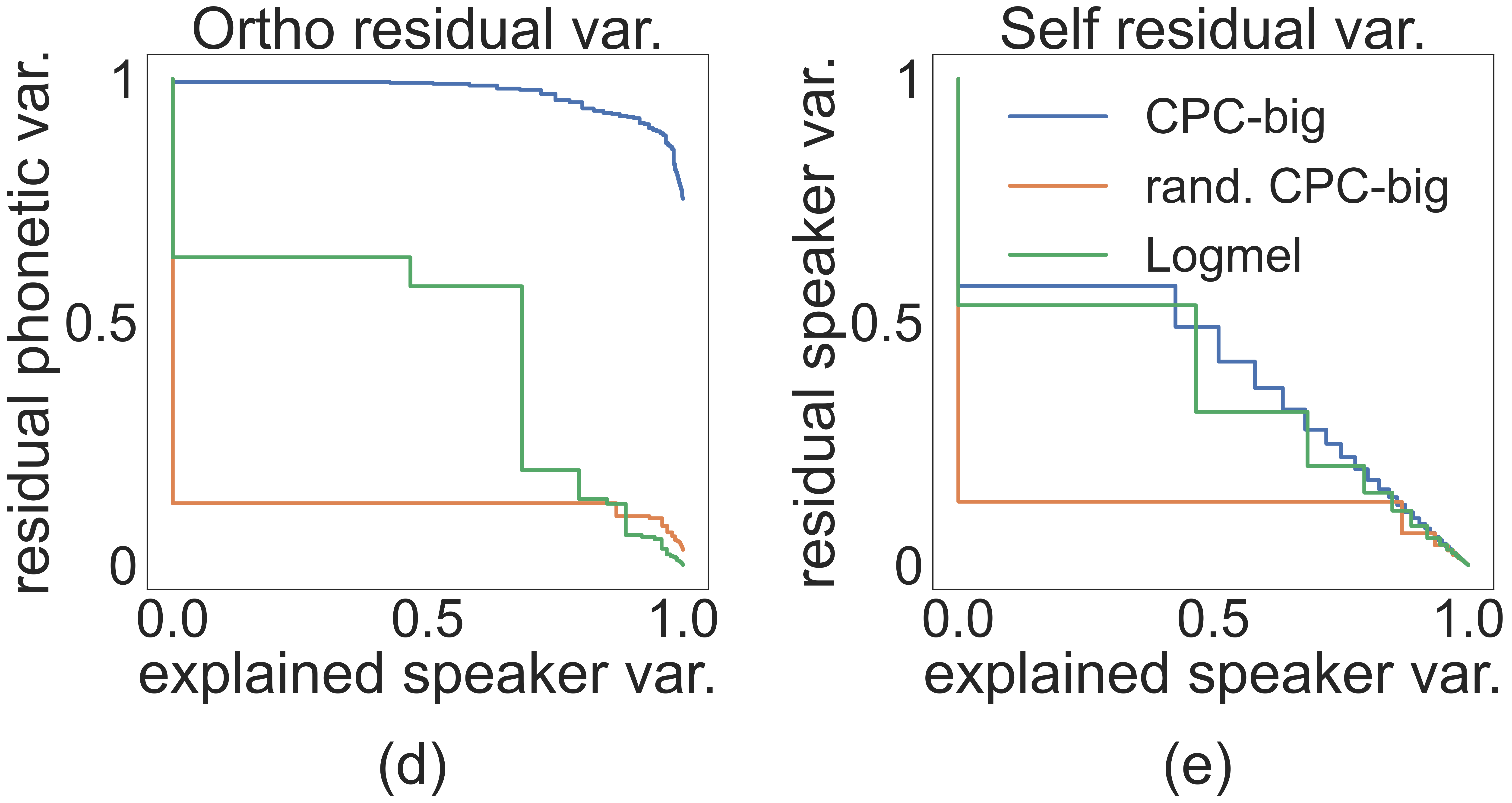}
    \vspace{-0.3cm}
    \caption{Evaluating orthogonality as cosine similarities between principal components (a-c) versus using residual variance (d). Isotropy can be evaluated with self residual variance (e).}
    \vspace{-0.4cm}
    \label{fig: similarity versus residual variance}
\end{figure*}

\vspace{-0.5em}

\section{Isotropy and orthogonality}

In NLP, most researchers have argued that representations with greater isotropy are desirable \cite{cai2020isotropy,mu2017all,timkey2021all,li2020sentence}; but see \cite{ding2021isotropy}. However, Rudman \textit{et al.}\ \cite{rudman2022isoscore} noted that the measures of ``isotropy" used in much of that work do not match its mathematical definition---that is, the extent to which the covariance matrix is proportional to the identity matrix. They introduced (and demonstrated the correctness of) a new measure called IsoScore, and later used it to show that isotropy is in fact {\em negatively} correlated with task performance in several BERT models \cite{rudman2023stable}.
Meanwhile, we know of only one study of isotropy in models of speech \cite{abdullah2023analyzing}, which found a strong positive correlation between IsoScore and word discrimination performance in supervised acoustic word embedding models.
Here, we explore whether isotropy can predict either phone or speaker classification performance in self-supervised representations.

As noted above, IsoScore \cite{rudman2022isoscore} is one way to measure isotropy. IsoScore ranges from 0 (minimally isotropic) to 1 (maximally isotropic), and can be interpreted as the approximate proportion of the dimensions that are uniformly utilized. 
Computing the IsoScore for a point cloud \(\data{X}  \subseteq \repspace\) starts by applying 
PCA, then finding the Euclidean distance between the length-normalized vector of eigenvalues $\Lambda$ (the diagonal of the covariance matrix) and the diagonal of the identity matrix in $\repspace$. This distance is then normalized and rescaled to fall between 0 and 1.\looseness=-1

While IsoScore has properties that can be desirable (e.g., it allows direct comparisons between spaces of different dimensionality on the same 0-1 scale), it is not the only possible measure of isotropy. For example, Del Giudice \cite{del2021effective} discusses estimators of ``Effective Dimensionality'' which normalize $\Lambda$ to create a probability distribution, then calculate its entropy to measure deviance from uniformity, and return a value interpreted as the {\em number} (rather than {\em proportion}) of dimensions uniformly utilized. 

Apart from isotropy, orthogonality is also a desirable property when learning representations, since encoding different kinds of information in orthogonal dimensions or subspaces would allow them to be easily disentangled.
In fact, there have been attempts in representation learning to enforce such orthogonality to enable disentanglement \cite{sarhan2020fairness, pmlr-v202-cha23b}.
There is also evidence that human brains encode different aspects of the same item in orthogonal coding axes, thereby minimizing interference and maximizing robustness \cite{flesch2022orthogonal,libby2021rotational}.
However, \cite{liu23_orthogonal} (henceforth, \ltg) is the only work we know of to explore orthogonality in either supervised or unsupervised speech models. We describe their method, and how we build on it, in more detail below.

\section{Measuring orthogonality}
\label{sec:measuring-ortho}
Before evaluating the orthogonality between speaker and phonetic encoding, we first follow \ltg\ in identifying speaker directions and phonetic directions.
Phonetic directions are found by aggregating the frame-level representations for each of the 39 phones (based on forced alignment) to obtain their centroids, and then applying principal component analysis to the centroids.
The 39 principal components found represent the \emph{phonetic directions}, along which the variance between the centroids is maximized.
The same method is used to obtain speaker directions using the speakers in the dataset.

Our next step diverges from \ltg's:
while they looked at the cosine similarities between the phonetic and speaker directions (\S\ref{sec:cosine-similarity}), we propose a new measure that quantifies orthogonality with a single numerical value (\S\ref{sec:crv}).

\subsection{Cosine similarity between principal directions}\label{sec:cosine-similarity}
For each pair of a speaker and a phonetic direction, \ltg\ computed the degree of orthogonality 
by taking the absolute value of their cosine similarity, with 0 being perfectly orthogonal and 1 being perfectly aligned.
Figs.~\ref{fig: similarity versus residual variance}a-c present the pairwise similarity for representations extracted from (a) the second LSTM layer of the same CPC-big model used by \ltg\ (from \cite{nguyen2020zero}); (b) the same layer of a randomly initialized CPC model that has not been trained, and (c) log Mel features.
Confirming \ltg's results, Fig.~\ref{fig: similarity versus residual variance}a shows very low similarities between any pair of phonetic and speaker directions, indicating the two types of information are largely encoded orthogonally.

While the similarity matrix gives some indication of the relationship between the directions encoding speaker and phonetic information, it can be difficult to summarize with a single number: we need to consider the degree of alignment between every pair of directions in order to fully capture the degree of orthogonality between speaker and phonetic encoding.
In addition, alignment between principal directions with large eigenvalues means overall lower orthogonality than alignment between principal directions with small eigenvalues,
but the matrix does not reflect the amounts of variance in each principal direction.

\subsection{Cumulative Residual Variance (CRV)}
\label{sec:crv}

We propose Cumulative Residual Variance (CRV) as a quantitative measure of orthogonality between datasets \data{X} and \data{Y} embedded in $\repspace$. CRV satisfies the two desiderata mentioned above: (1) it captures the interaction between every pair of principal directions and (2) it weights the contribution from each principal direction in proportion to its relative explained variance.
Here, we set \data{X} and \data{Y} to be the speaker and phone centroids (or vice versa), so the number of data points $n_X$ and $n_Y$ is less than the dimensionality $d$, and each dataset only spans a subspace of \repspace. However, this need not be true in general; for example CRV could be applied to the sets of frame-level representations from two different speakers or two different phones.

In short, the CRV of \data{Y} with respect to \data{X}, written as \crv{\data{Y}}{\data{X}}, evaluates how much variance is preserved in \data{Y} as the principal directions of \data{X} are collapsed one by one.\footnote{Note that CRV is an asymmetrical distance measure. Like KL divergence, it could be symmetrised as \crv{\data{X}}{\data{Y}} + \crv{\data{Y}}{\data{X}}, if desired.
}
As in \ltg, ``collapsing" a direction $v$ from a dataset $\data{Y}$ refers to the operation of projecting \data{Y} onto the subspace orthogonal to $v$, i.e. $\data{Y}' = \data{Y} - (\data{Y}v)v^\top$.
Collapsing $v$ affects any principal direction of \data{Y} that is not orthogonal to $v$, which addresses the first desideratum.
We evaluate the effect of the collapsing operation by computing the residual variance in $\data{Y}'$, as given by PCA.
The larger the residual variance \data{Y'} has, the more orthogonal \data{Y} is to $v$.

The residual variances computed in this way can be plotted as in Fig.\ \ref{fig: similarity versus residual variance}d, where for any given $x$-axis value, its $y$ value is the proportion of variance remaining in \data{Y} after collapsing the minimum number of top principal directions of \data{X} such that at least $x$ proportion of \data{X}'s variance has been removed.
CRV is then computed from this plot as the area under the curve (AUC), to yield a single numerical value. 
In this way, the effect of collapsing each direction is weighted by the variance explained by that direction, hence CRV also satisfies our second desideratum. 

In Fig.~\ref{fig: similarity versus residual variance}d, we plot residual variance of the phone centroids with respect to the explained variance in the speaker centroids for representations from a trained and an untrained CPC\footnote{Though CPC is a loss function, with a slight abuse, we refer to a randomly initialized CPC-big in \cite{nguyen2020zero} as untrained CPC.} as well as for log Mel features.
We can see that the relative magnitude of the AUC is CPC, followed by log Mel and untrained CPC.
While the strong orthogonality in the trained CPC is consistent with Fig.\ \ref{fig: similarity versus residual variance}a, the relative degree of orthogonality between log Mel and untrained CPC is less salient from Fig.\ \ref{fig: similarity versus residual variance}b-c. 
There are more dark spots in Fig.\ \ref{fig: similarity versus residual variance}c, indicating more pairs of aligned speaker and phonetic directions in log Mel, but this should have less effect on overall orthogonality as compared to the top left corner of Fig.\ \ref{fig: similarity versus residual variance}b, which shows that the first two speaker and phonetic directions of the untrained CPC are very strongly aligned.
This is properly reflected in Fig.\ \ref{fig: similarity versus residual variance}d.

\vspace{-0.5mm}

\subsection{Evaluating isotropy with Self-CRV}
A byproduct of CRV is \emph{self-CRV},
or \crv{\data{Y}}{\data{Y}}, which evaluates the degree of isotropy of \data{Y} in \repspace.
If \data{Y} is highly anisotropic, its variance will be concentrated around a few directions. 
This results in a residual variance curve with a small AUC,
as illustrated in Fig.~\ref{fig: similarity versus residual variance}e for untrained CPC.

Self-CRV is closely related to IsoScore, both being functions of the eigenvalues of the dataset.
However, IsoScore measures isotropy as a percentage of representation dimensions, whereas self-CRV accounts for the absolute number of isotropic dimensions and is comparable across models with different dimensions
as long as the number of data points in \data{Y} is smaller than the dimensionality of all models (as in our subspace analyses).
After multiplying IsoScore by model dimension, we found a Spearman's rank correlation of 1 between it and self-CRV.

\vspace{-1mm}

\section{Experimental Setup}

\paragraph*{Models}
In addition to CPC-big, we measured orthogonality and isotropy in five pre-trained Transformer-based English self-supervised speech models: HuBERT (base-ls960) \cite{hsu_hubert_2021}, wav2vec 2.0 (base-960h) \cite{baevski2020wav2vec}, WavLM (base) \cite{chen2022wavlm}, WavLM+ (base-plus) \cite{chen2022wavlm}, and Data2Vec (base-960h) \cite{baevski2022data2vec}.
Apart from the architecture, CPC-big differs from the Transformer-based models in its dimensionality (512 vs.\ 768), number of layers (5 CNN followed by 4 LSTM vs.\ 7 CNN followed by 12 Transformer blocks), frame rate (10ms vs.\ 20ms) and amount of training data (6k hr vs.\ 960 hr for all others except WavLM+, which used 96k hr).
To determine the degree of orthogonality and isotropy in these models before training, 
we also tested representations extracted from a HuBERT model and a CPC-big model with just random initialization and no training. 
Since the Transformer models we tested have mostly the same architecture and are distinguished by the training methods and objective, the untrained HuBERT is representative of the other Transformer models. 
Finally, 40-dimensional log Mel features are used as a  baseline.

\vspace{\myparspace}\paragraph*{Dataset}
We perform our analysis on the dev-clean subset of LibriSpeech \cite{panayotov2015librispeech}, which matches the language (English) and genre (read speech) of the pre-trained models and was also used in \ltg\footnote{We hope in future to examine how much these results generalize, by extending the analyses 
to other genres and languages, either using different pre-trained models or by testing these models on other data.}
Dev-clean contains 40 speakers, each contributing at least eight minutes of speech. We used half of dev-clean for training classifiers and half for testing, with different splits depending on the scenario, as described below.

\vspace{\myparspace}\paragraph*{Probing classifiers}
\label{sec:classifier setup}
Our analysis focuses on speaker information and phonetic information, due to their influence on a variety of downstream speech tasks. 
We train logistic regression classifiers to predict the speaker (or phone) label based on a single representation frame. In previous work, frames are typically pooled across phones \cite{pasad2023comparative,english2022domain} or utterances (for speaker ID) \cite{yang21c_interspeech}; but like \ltg\ we use individual frames, so we can analyze how both types of information sit in the same set of embeddings.\looseness=-1

For speaker classification, we obtain speaker labels from the LibriSpeech metadata and train the probing classifier on a random half of each speaker’s utterances, using the other half for testing. 
For phone classification, we obtain the phone labels from forced alignments with Kaldi.
We evaluated phone accuracy in two ways: {\em shared speakers} (as in \ltg), where the same speakers appear in both training and test, and the more standard {\em across-speaker}, where we trained on data from a random half of the speakers and tested on the other half. In practice, the measures are very strongly correlated and don't differ much, so in this paper we only report across-speaker phone accuracy.

\vspace{\myparspace}\paragraph*{Computing CRV, IsoScore, and correlations}

We computed CRV and IsoScore for each layer of each model by first encoding the utterances from LibriSpeech dev-clean to obtain the representations. We then computed the phone and speaker centroids and CRV values as described in \S\ref{sec:measuring-ortho}. In particular, \crv{Ph}{Spk} measures the orthogonality of the phone and speaker subspaces, and \crv{Ph}{Ph} and \crv{Spk}{Spk} measure the isotropy of the phone and speaker subspaces, respectively. We computed the IsoScore using a random sample of 250,000 frames. Finally, for correlations between classifier accuracy and CRV or IsoScore, we computed Spearman (rank) correlation, since it is less sensitive to outliers and we have no reason to believe that correlations will be linear.

\vspace{-1mm}

\section{Results and Discussion}


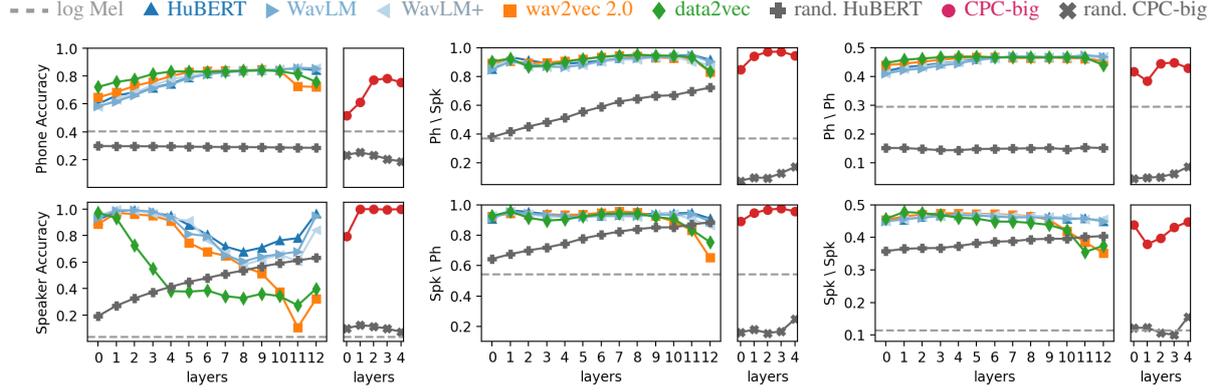
\begin{figure*}[t]
    \definecolor{logmel}{HTML}{999999}
    \definecolor{hubert}{HTML}{1F77B4}
    \definecolor{wavlm}{HTML}{78ADD2}
    \definecolor{wavlmplus}{HTML}{BBD6E8}
    \definecolor{wav2vec}{HTML}{FF7F0E}
    \definecolor{data2vec}{HTML}{2CA02C}
    \definecolor{rand}{HTML}{666666}
    \definecolor{cpc}{HTML}{D62768}
    
    \begin{center}
    \begin{tikzpicture}[font=\footnotesize]
    \draw[dashed, logmel, ultra thick] (0, 0) -- (0.5, 0);
    \node[right, logmel] at (0.5, 0) {log Mel \strut};
    \end{tikzpicture}
    \begin{tikzpicture}[font=\footnotesize]
    \node[regular polygon, regular polygon sides=3, inner sep=1.2pt, fill=hubert] at (0.4, 0) {};
    \node[right, hubert] at (0.5, 0) {HuBERT \strut};
    \end{tikzpicture}
    \begin{tikzpicture}[font=\footnotesize]
    \node[regular polygon, regular polygon sides=3, rotate=-90, inner sep=1.2pt, fill=wavlm] at (0.4, 0) {};
    \node[right, wavlm] at (0.5, 0) {WavLM \strut};
    \end{tikzpicture}
    \begin{tikzpicture}[font=\footnotesize]
    \node[regular polygon, regular polygon sides=3, rotate=90, inner sep=1.2pt, fill=wavlmplus] at (0.4, 0) {};
    \node[right, wavlmplus!75!black] at (0.5, 0) {WavLM+ \strut};
    \end{tikzpicture}
    \begin{tikzpicture}[font=\footnotesize]
    \node[regular polygon, regular polygon sides=4, inner sep=1.7pt, fill=wav2vec] at (0.4, 0) {};
    \node[right, wav2vec] at (0.5, 0) {wav2vec 2.0 \strut};
    \end{tikzpicture}
    \begin{tikzpicture}[font=\footnotesize]
    \node[diamond, aspect=0.75, inner sep=1.5pt, fill=data2vec] at (0.4, 0) {};
    \node[right, data2vec] at (0.5, 0) {data2vec \strut};
    \end{tikzpicture}
    \begin{tikzpicture}[font=\footnotesize]
    \node[rectangle, inner sep=1.2pt, fill=rand, minimum height=6pt] at (0.4, 0) {};
    \node[rectangle, inner sep=1.2pt, fill=rand, minimum width=6pt] at (0.4, 0) {};
    \node[right, rand] at (0.5, 0) {rand.\ HuBERT \strut};
    \end{tikzpicture}
    \begin{tikzpicture}[font=\footnotesize]
    \node[circle, inner sep=1.8pt, fill=cpc] at (0.4, 0) {};
    \node[right, cpc] at (0.5, 0) {CPC-big \strut};
    \end{tikzpicture}
    \begin{tikzpicture}[font=\footnotesize]
    \node[rectangle, inner sep=1.2pt, rotate=45, fill=rand, minimum height=7pt] at (0.4, 0) {};
    \node[rectangle, inner sep=1.2pt, rotate=-45, fill=rand, minimum height=7pt] at (0.4, 0) {};
    \node[right, rand] at (0.5, 0) {rand.\ CPC-big \strut};
    \end{tikzpicture}
    
    \includegraphics[width=5.1cm]{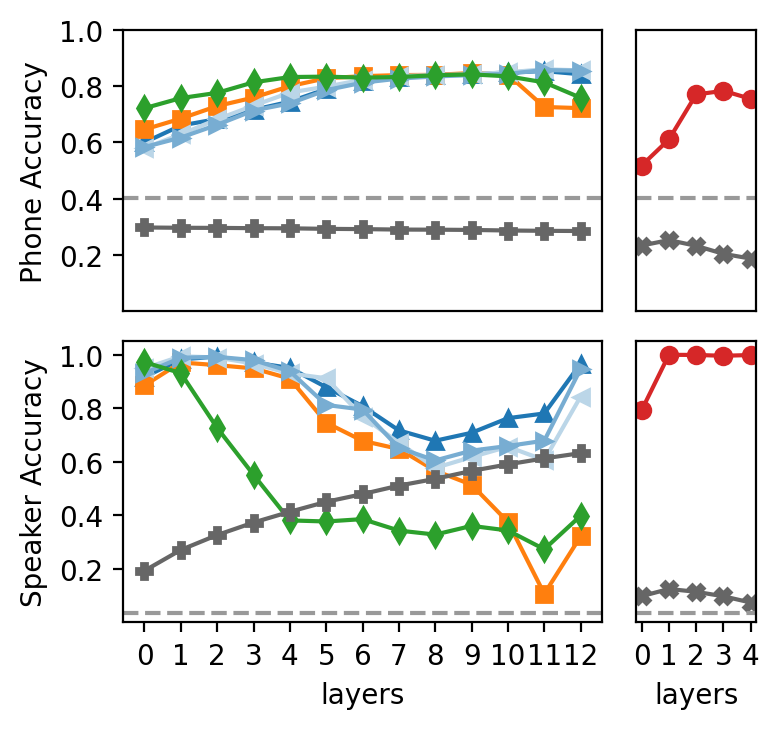}
    \includegraphics[width=5.1cm]{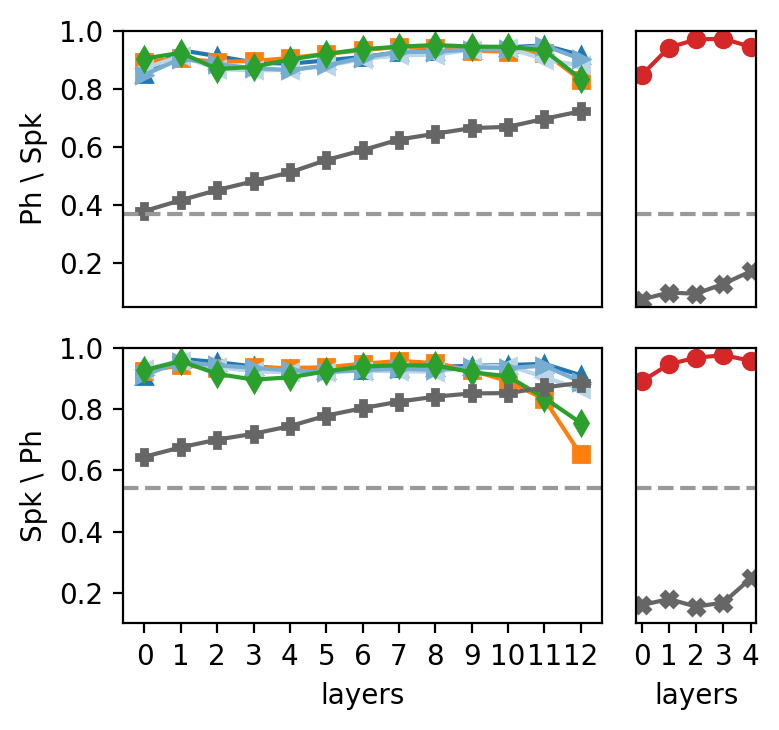}
    \includegraphics[width=5.1cm]{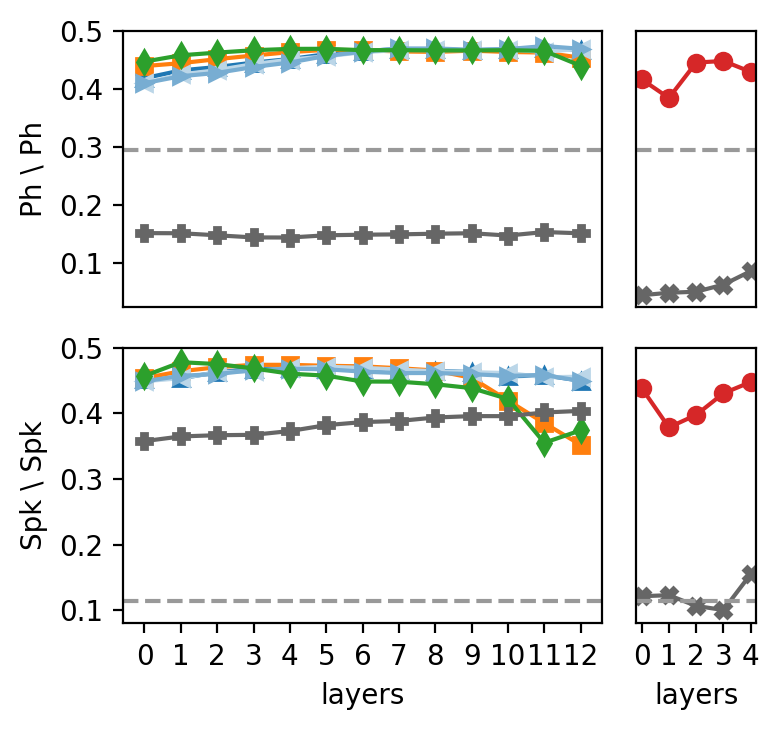}
    \end{center}
    \vspace{-7mm}
    \caption{Layerwise results for all models, showing (in columns from left to right): Phone and speaker classiciation accuracy; CRV orthogonality measures (\crv{Ph}{Spk} and \crv{Spk}{Ph}); and self-CRV measures (\crv{Ph}{Ph} and \crv{Spk}{Spk}). }
    \vspace{-0.5em}
\label{fig:AUC SPK vs PH}
\label{fig:AUC PH vs SPK}
\label{fig:phone and speaker ACC}

\end{figure*}


\begin{figure*}
    \centering
    \includegraphics[height=3.2cm]{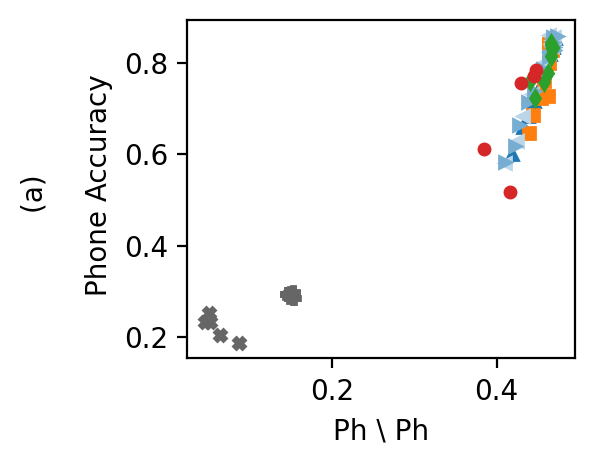}
    \includegraphics[height=3.2cm]{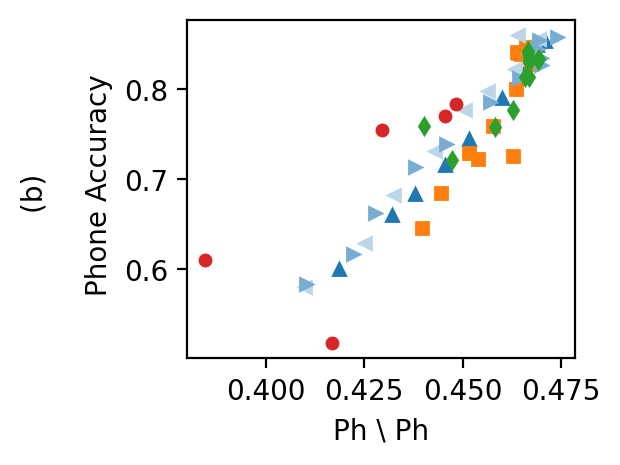}
    \includegraphics[height=3.2cm]{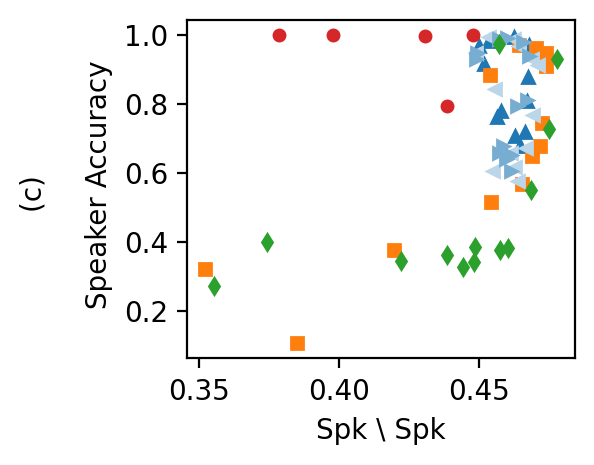}
    \includegraphics[height=3.2cm]{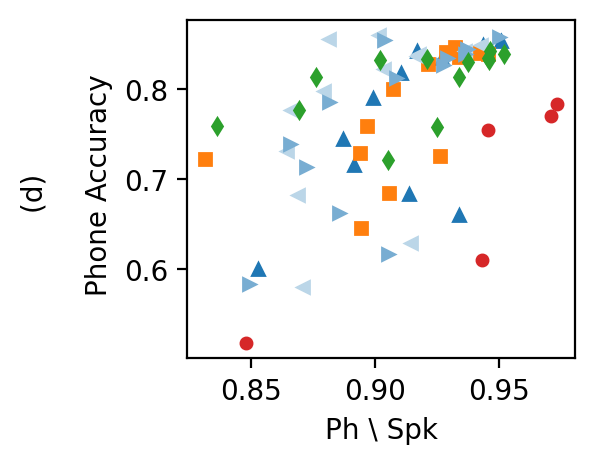}
    \vspace{-2mm}
    \caption{Correlations between orthogonality or isotropy measures and probing accuracies. Marker styles are as in Fig.~\ref{fig:phone and speaker ACC}. 
    }
    \vspace{-1.2em}
 \label{fig:PhXPhvsPhAcc(XS)}
\end{figure*}

\subsection{Layerwise Classification Accuracy}

Fig.~\ref{fig:phone and speaker ACC} (lst column) shows the results of our probing classifiers for phones (top) and speakers (bottom), across all layers of each model\footnote{Due to space constraints, not all results can be displayed in the paper. The complete spreadsheet of our results, and the code for computing CRV can be found at \url{https://github.com/uililo/cumulative-residual-variance}.}.
For phones, our findings align closely with those of \cite{pasad2021layer, pasad2023comparative,english2022domain}, despite analyzing frame-wise rather than pooling the representations for each phone token.
That is, for wav2vec2 (and data2vec) the highest probing accuracies are in the late middle layers, while for HuBERT-family models (HuBERT, WavLM, WavLM+), accuracy remains high through the final layers.

To the best of our knowledge, previous studies have only reported speaker probing accuracy across all the layers of HuBERT \cite{chang23_interspeech}.
Extending the layerwise analysis of speaker information to the other widely-used SSL models, we find far more variation here than with phone accuracy, perhaps because all of these models, despite being self-supervised, are designed with ASR in mind.
We see especially poor linear separability in the later layers of wav2vec2 and data2vec, where speaker accuracy is even worse than the randomly initialized ones.
We speculate that the rising pattern of speaker accuracy in the randomly initialized models may be because the model incorporates more context in later layers, allowing the model to effectively average features over the whole utterance.

\vspace{-0.5mm}

\subsection{Geometry of the phone and speaker subspaces}
Layer-wise CRV and self-CRV results for all models are shown in Fig.~\ref{fig:AUC PH vs SPK}, columns 2 and 3. Like the CPC model studied by \ltg, all trained Transformer models have high \crv{Ph}{Spk} orthogonality. Interestingly, untrained HuBERT (unlike untrained CPC) also reaches a somewhat high \crv{Ph}{Spk} value in the final layers, although still lower than the trained models. The trained models also show high isotropy in the phone and speaker centroids (\crv{Ph}{Ph} and \crv{Spk}{Spk}), though as with probing accuracy, the difference between trained and untrained models is much more striking for the phonetic measure, suggesting that model training reorganizes the representational geometry of the phonetic information more than the speaker information.

We then computed rank correlations $\rho$ between each of the four CRV measures and the speaker or phone accuracies.
The most striking correlation is between \crv{Ph}{Ph} and phone accuracy, as shown in Figs.~\ref{fig:PhXPhvsPhAcc(XS)}a (all models) and~\ref{fig:PhXPhvsPhAcc(XS)}b (trained models only). Pooling all datapoints together, $\rho$ = 0.94, and the trained models individually each have $\rho$ from 0.69 to 0.9 (all values $p<0.05$). In contrast, we only found statistically significant correlations between \crv{Spk}{Spk} and speaker accuracy (Fig.~\ref{fig:PhXPhvsPhAcc(XS)}c) in wav2vec2 and data2vec, and no significant correlation when pooling the results from all trained models.

For the orthogonality measures, we found significant correlations between \crv{Ph}{Spk} and phone accuracy (Fig.~\ref{fig:PhXPhvsPhAcc(XS)}d) in each of the trained models ($\rho=$ 0.54-0.78 for Transformer models, 1.0 for CPC), as well as in the pooled data ($\rho=$ 0.54), although the correlations are weaker than for \crv{Ph}{Ph}. Correlations between \crv{Spk}{Ph} and speaker accuracy are even weaker, reaching significance on the pooled data, but not for any individual model. 

Altogether, our results suport \ltg's claim that orthogonality between the phonetic and speaker subspace is relevant for extracting phonetic information, but also suggest that isotropy of the phonetic space may be even more critical. It is less clear why the geometry of speaker information is less correlated to speaker classification, and to what extent this result is due to model training that is implicitly focused on ASR performance.

\subsection{Isotropy of the frame representation space}
Finally, we evaluated the isotropy of frame representations (rather than the centroids).
For this, we used IsoScore, which (1) has a rank correlation of 1 with self-CRV as isotropy measures of speaker or phone centroids, and (2) is easier to compute than self-CRV when applied to a large number of representations. 
The IsoScore values were low, ranging from 0.18 to near 0 across models and layers, similar to the range found by Rudman  \textit{et al.}\  \cite{rudman2022isoscore} for contextualized word embedding models.
Also, the IsoScore values for untrained HuBERT were comparable to those of the trained models. 
We find a statistically significant ($p<0.05$) positive correlation with phone probing accuracy in HuBERT and WavLM, and when pooling results from all trained models; but for speaker probing accuracy we found negative correlations in the same two models, and no significant pooled correlation. 
These mixed results suggest that isotropy of the representation space itself is not necessarily a good predictor of model performance, especially if different tasks are considered. 

\section{Conclusion}

This paper introduced the Cumulative Residual Variance as a new way to analyze the representational geometry of high-dimensional spaces, and used CRV and IsoScore to examine whether orthogonality or isotropy can predict phone or speaker probing accuracy in self-supervised speech models. We did not find strong evidence that isotropy of the frame representations is meaningful, but we did show that phone probing accuracy is correlated with the degree of orthogonality between the subspaces defined by the phone and speaker centroids, and even more strongly with the isotropy of the phone centroids themselves. These findings suggest that geometric analyses may be a productive route for future study, 
particularly if they can be more closely connected to theoretical analyses such as those of \cite{sorscher2022neural}.  
For instance, \cite{sorscher2022neural} highlights the relevance of four different geometric properties, including the distance between class centroids (related to our subspace isotropy measure) as well as the isotropy of the individual class manifolds (i.e., phones or speakers). We hope that our work may inspire further exploration of these connections. 

\section{Acknowledgements}
This work was supported in part by the UKRI Centre for Doctoral Training in Natural Language Processing, funded by the UKRI (grant EP/S022481/1)
and the University of Edinburgh, and also received support from Google.
\bibliographystyle{IEEEtran}
\bibliography{mybib,references_sgwater}

\end{document}